\documentclass[11pt]{article}

\usepackage[letterpaper,margin=1in]{geometry}
\usepackage{amsmath,amssymb}
\usepackage{graphicx}
\usepackage{booktabs}
\usepackage{cite}
\usepackage{multirow}
\usepackage{xcolor}
\usepackage{array}
\usepackage[hidelinks]{hyperref}

\begin{document}

\title{Baseline-Free Policy Optimization for\\Neural Combinatorial Optimization}

\author{
Carlos S. Sep\'ulveda\\
Facultad de Ingenier\'ia y Ciencias, Universidad Adolfo Ib\'a\~nez\\
Direcci\'on de Programas, Investigaci\'on y Desarrollo, Armada de Chile\\
Valpara\'iso, Chile\\
\texttt{carlos.sepulveda@alumnos.uai.cl}
\and
Gonzalo A. Ruz\\
Facultad de Ingenier\'ia y Ciencias, Universidad Adolfo Ib\'a\~nez\\
Millennium Nucleus for Social Data Science (SODAS)\\
Millennium Nucleus in Data Science for Plant Resilience (PhytoLearning)\\
Santiago, Chile\\
\texttt{gonzalo.ruz@uai.cl}
}

\date{Preprint, June 2026}

\maketitle

\begin{abstract}
Neural combinatorial optimization (NCO) trains autoregressive
policies to solve routing problems. The standard training algorithm, REINFORCE with a rollout baseline, requires maintaining and periodically updating a frozen copy of the policy for variance reduction. This baseline introduces a structural vulnerability: on harder instances, a poor  baseline produces noisy gradient estimates that can destabilize training. We evaluate Group Relative Policy
Optimization (GRPO), an algorithm from large language model alignment that eliminates the baseline entirely by normalizing advantages within groups of sampled trajectories. In a controlled comparison of five RL algorithms on TSP and CVRP benchmarks within the RL4CO framework, we find that: (i)~GRPO avoids the training collapse observed with REINFORCE on TSP-100, where performance degrades from cost 9.8 to 52.1 immediately after the warmup phase and does not recover under extended training; (ii)~at matched gradient updates, GRPO achieves solution quality within 2\% of POMO, a strong AM-based multi-start baseline, while requiring no external baseline; and (iii)~P3O, a pairwise preference algorithm also from the alignment literature, is competitive on TSP but shows higher variability on CVRP. These results identify GRPO as a promising baseline-free alternative for NCO, particularly in settings where baseline-dependent training becomes fragile.
\end{abstract}


% ==============================================================
\section{Introduction}\label{sec:intro}
% ==============================================================

Neural combinatorial optimization (NCO) uses autoregressive neural networks to construct solutions for NP-hard routing problems~\cite{Kool2018,berto2025rl4co}. The dominant training paradigm is REINFORCE~\cite{Williams1992} with a rollout baseline~\cite{Kool2018}: a frozen copy of the current policy generates greedy solutions that serve as a reference for computing advantages. POMO~\cite{kwon2020pomo} extended this with multiple starting points and a shared baseline. Both methods rely on an external baseline mechanism for variance reduction.

This dependence on a baseline introduces a vulnerability. On small instances (TSP-50), any reasonable policy produces acceptable greedy tours, so the rollout baseline provides a useful reference. On larger or more constrained instances, the baseline quality degrades alongside the policy quality, creating a circular dependency that can destabilize training. We demonstrate this failure concretely: on TSP-100, REINFORCE with rollout baseline learns during a brief warmup phase (cost drops from 52 to 9.8 in one epoch) but degrades catastrophically once the rollout baseline activates, never recovering even after 800 epochs.

Group Relative Policy Optimization (GRPO)~\cite{shao2024deepseekmath},
originally proposed for aligning large language models (LLMs), offers a principled alternative. GRPO generates $G$ trajectories per instance and computes advantages as z-scores within each group, requiring no external baseline. This paper investigates whether this baseline-free approach can benefit NCO.

We evaluate GRPO, P3O~\cite{Wu2023}, PPO~\cite{Schulman2017},
REINFORCE~\cite{Kool2018}, and POMO~\cite{kwon2020pomo} on four standard benchmarks using the RL4CO framework~\cite{berto2025rl4co}.
To ensure a fair comparison, we report results at both equal training epochs and equal gradient updates, since GRPO's multi-epoch inner loop performs approximately $8\times$ more gradient updates per training epoch than single-update methods.

Our contributions are:
\begin{enumerate}
\item To the best of our knowledge, the first evaluation of GRPO and P3O on standard NCO benchmarks, with both epoch-matched and gradient-update-matched comparisons.
\item Evidence that REINFORCE with a rollout baseline exhibits a reproducible failure mode on TSP-100 in the tested RL4CO configuration, persisting across two seeds and extended training.
\item A finding that, at matched gradient updates, GRPO achieves competitive quality with POMO-4 (within 2\%) while using no external baseline, establishing it as a viable baseline-free alternative for NCO.
\end{enumerate}

% ==============================================================
\section{Related Work}\label{sec:related}
% ==============================================================

\textbf{Neural combinatorial optimization.}
Recent surveys further frame graph reinforcement learning as a unifying perspective for combinatorial optimization~\cite{darvariu2024graph}.
Pointer Networks~\cite{Vinyals2015} first applied attention-based sequence models to TSP, and Bello et al.~\cite{Bello2016} introduced policy-gradient training for NCO. Subsequent work extended neural routing to VRP variants~\cite{Nazari2018,delarue2020reinforcement,Li2021a,Xin2021}. The Attention Model (AM)~\cite{Kool2018} combined Transformer-style encoders with a rollout baseline and became a standard architecture for learned routing. POMO~\cite{kwon2020pomo} improved AM-style training by exploiting multiple equivalent optima through multi-start rollouts. RL4CO~\cite{berto2025rl4co} provides a unified implementation and benchmark suite for these methods.

\textbf{Robotics and automation relevance.}
TSP, VRP, and coverage variants are canonical abstractions behind mobile-robot inspection, UAV/ASV surveillance, agricultural coverage, and fleet-level task allocation. Coverage path planning is a central robotics problem~\cite{Galceran2013}. Recent robotic surveillance, coverage, and learning-based planning studies further illustrate that practical planning pipelines often include waypoint ordering, route construction, or constrained routing
components~\cite{Faigl2018,Karapetyan2019,Mier2023,Shah2020,jonnarth2024learning}.

\textbf{RL baselines for NCO.}
Baseline design is critical for policy-gradient NCO. AM uses a rollout baseline from a frozen policy snapshot~\cite{Kool2018}; POMO uses a shared baseline across multiple starts~\cite{kwon2020pomo}; PPO replaces explicit rollout baselines with a learned critic~\cite{Schulman2017}. These choices affect variance reduction, sample reuse, and stability.

\textbf{Relative and preference-based policy optimization.}
GRPO~\cite{shao2024deepseekmath} and P3O~\cite{Wu2023} were introduced in LLM alignment, using group-relative or pairwise comparisons rather than learned value functions. Concurrently, Pan et al.~\cite{pan2025po4cops} proposed preference optimization for combinatorial optimization fine-tuning. Our work differs by evaluating GRPO and P3O as direct training methods for standard learned-routing benchmarks.

% ==============================================================
\section{Methods}\label{sec:methods}
% ==============================================================

\subsection{Problem Setting}

We consider TSP and CVRP. An autoregressive policy $\pi_\theta(a_t|s_t)$ constructs a feasible solution $\tau$ by selecting nodes sequentially. Let $C(\tau)$ denote the solution cost: the tour length for TSP and the total route length for CVRP. Since the learning algorithms are formulated as reward maximization, we define $R(\tau)=-C(\tau)$ and report $C(\tau)$ in all tables, where lower is better.

\subsection{REINFORCE with Rollout Baseline}

REINFORCE~\cite{Williams1992,Kool2018} maximizes expected reward
$J(\theta)=\mathbb{E}_{\tau\sim\pi_\theta}[R(\tau)]$ using
\[
\nabla_\theta J =
\mathbb{E}\left[
\left(R(\tau)-b_R(\mathbf{x})\right)
\nabla_\theta \log \pi_\theta(\tau|\mathbf{x})
\right],
\]
where $b_R(\mathbf{x})=-C_{\mathrm{greedy}}(\mathbf{x})$ is the reward equivalent of the greedy-rollout cost produced by a frozen policy.
Equivalently, in cost-minimization form, the advantage is
$C_{\mathrm{greedy}}(\mathbf{x})-C(\tau)$. In RL4CO, this is wrapped in a \texttt{WarmupBaseline}~\cite{Kool2018} that returns a convex
combination of an exponential moving average baseline and the rollout baseline during a warmup phase (\texttt{n\_epochs}=1 by default), transitioning to the pure rollout baseline afterward.
POMO~\cite{kwon2020pomo} generates $G$ solutions from different
starting nodes and uses a shared baseline (mean reward across starts).

\subsection{GRPO for Combinatorial Optimization}

For each instance in a batch, GRPO generates $G$ trajectories by
sampling from $\pi_\theta$ with temperature $T$ and computes
intra-group advantages:
\begin{equation}\label{eq:grpo_adv}
\hat{A}_g = \frac{R(\tau_g) - \mu_G}{\max(\sigma_G,\, \epsilon)}
\end{equation}
where $\mu_G$ and $\sigma_G$ are the mean and standard deviation of
the $G$ returns. 
Because $R(\tau)=-C(\tau)$, lower-cost trajectories receive higher
relative rewards and therefore positive normalized advantages when they outperform the group mean.
The policy is updated using PPO-clip:
\begin{equation}\label{eq:grpo_loss}
\mathcal{L} = -\frac{1}{BG}\sum_{b,g}
\min\!\bigl(r_{bg}\hat{A}_{bg},\;
\operatorname{clip}(r_{bg}, 1{\pm}\varepsilon)\hat{A}_{bg}\bigr)
\end{equation}
where $B$ is the batch size, $G$ is the group size, and
$r_{bg}=\pi_\theta(\tau_{bg}|\mathbf{x}_b)/
\pi_{\theta_{\mathrm{old}}}(\tau_{bg}|\mathbf{x}_b)$ is the trajectory probability ratio. After collecting the $BG$ trajectories, we apply Eq.~\eqref{eq:grpo_loss} for $K$ inner PPO epochs using mini-batches
drawn from the stored trajectories.

Two properties distinguish GRPO from baseline-dependent methods.
First, the advantage computation is self-contained: no frozen policy copy or critic is needed. Second, group-relative normalization makes the update invariant to affine reward scaling within each instance, because advantages are computed from the relative ranking of sampled trajectories rather than from their absolute cost scale.

A practical requirement for group-based methods is trajectory
diversity within each group: if all $G$ rollouts yield identical
tours (likely at low temperature for easy instances), the standard
deviation $\sigma_G$ approaches zero and advantages become
degenerate. To maintain diversity throughout training, we linearly
anneal the sampling temperature from $T_\text{init}{=}2.0$ to
$T_\text{final}{=}1.0$ over the first 30\% of training epochs. The
high initial temperature ensures diverse group members with
well-defined z-scores during early training; as training progresses, the temperature is reduced to move the sampling
distribution closer to the greedy evaluation regime while preserving early exploration.

\subsection{P3O for Combinatorial Optimization}

P3O~\cite{Wu2023} uses pairwise trajectory comparisons. For each
instance, two trajectories $(\tau_1, \tau_2)$ are sampled from
$\pi_\theta$ and the pairwise advantage is their reward difference:
$A_\text{pair} = R(\tau_1) - R(\tau_2)$.
Under the cost convention, this is equivalent to
$A_\text{pair}=C(\tau_2)-C(\tau_1)$, so the first trajectory receives a positive pairwise advantage when it has lower cost than the second.
The policy is updated by clipping the log-ratio of selection
probabilities around its old value:
\begin{equation}\label{eq:p3o_loss}
\mathcal{L}_\text{P3O} = -\mathbb{E}\bigl[\min\bigl(
A_\text{pair}\,\delta,\;
A_\text{pair}\,\text{clip}(\delta,\,
\delta_\text{old}{\pm}\varepsilon)\bigr)\bigr]
\end{equation}
where $\delta = \log\pi_\theta(\tau_1) - \log\pi_\theta(\tau_2)$ is
the current log-ratio and $\delta_\text{old}$ is its value under the previous policy. Like GRPO, P3O requires no external baseline, the pairwise comparison is self-referencing. We use the v2 clipping variant~\cite{Wu2023} and the same temperature schedule as GRPO.

\subsection{Compute Cost Analysis}\label{sec:compute}

GRPO, PPO, and P3O perform $K$ inner PPO epochs per training step,
each iterating over mini-batches, yielding approximately
$8$ gradient updates per outer batch (with $K{=}2$ and mini-batch
fraction $0.25$). REINFORCE and POMO perform one gradient update per
batch. Over 100 training epochs (391 batches each), this results in
$\sim$312K gradient updates for GRPO versus $\sim$39K for REINFORCE
and POMO, an $8\times$ ratio. We address this disparity explicitly through gradient-update-matched experiments (Section~\ref{sec:compute_results}).

% ==============================================================
\section{Experimental Setup}\label{sec:setup}
% ==============================================================

\subsection{Benchmarks and Architecture}

We evaluate on TSP-50, TSP-100, CVRP-50, and CVRP-100 with node
coordinates sampled uniformly in $[0,1]^2$. All five algorithms share the same Attention Model policy~\cite{Kool2018} built on a Transformer encoder~\cite{vaswani2017attention}: 6 layers, 8 attention heads, embedding dimension 128.

\subsection{Training Protocol}

All experiments use RL4CO~\cite{berto2025rl4co} with batch size 256, learning rate $10^{-4}$ (Adam), 100K on-the-fly training instances per epoch, mixed precision (FP16), and gradient clipping at norm 1.0.
Each configuration is run with two random seeds (42, 123).

Algorithm-specific settings: GRPO uses $G{=}4$, $K{=}2$ inner epochs, clip $\varepsilon{=}0.2$. P3O uses the same clip, inner epochs, and temperature annealing. REINFORCE uses the RL4CO rollout baseline with one warmup epoch. POMO uses
$G{=}4$ starts with shared baseline. PPO uses $\varepsilon{=}0.2$, $K{=}2$. GRPO and P3O anneal temperature $T{:}2{\to}1$ over 30 epochs; REINFORCE and POMO use $T{=}1$.

RL4CO generates fresh training instances on-the-fly each epoch;
no static dataset is reused. Over 100 epochs, each method sees
$100 \times 100\text{K} = 10$M unique instances. Extended
runs (800 epochs) see 80M unique instances proportionally. All experiments were conducted on a single NVIDIA RTX PRO 6000 Blackwell Max-Q GPU.

\subsection{Evaluation and Validation}

The best checkpoint is selected on a fixed 10K-instance validation
set, then evaluated on a separate fixed 10K-instance test set using
greedy decoding. All 40 checkpoints were independently validated
by reloading and re-evaluating, yielding $<$1.2\% deviation from
reported values (0 mismatches above 2\%).

% ==============================================================
\section{Results}\label{sec:results}
% ==============================================================

\subsection{Epoch-Matched Comparison}\label{sec:epoch_results}

\begin{table}[t]
\centering
\caption{Tour cost at 100 epochs (mean $\pm$ std over 2 seeds,
10K greedy). Lower is better. Best per column in \textbf{bold}.
All methods use the same architecture and data. Note: GRPO, P3O,
and PPO perform ${\sim}8\times$ more gradient updates per epoch
than REINFORCE and POMO (see Section~\ref{sec:compute}).}
\label{tab:epoch}
\small
\setlength{\tabcolsep}{4pt}
\begin{tabular}{lcccc}
\toprule
& TSP-50 & TSP-100 & CVRP-50 & CVRP-100 \\
\midrule
GRPO
& $\mathbf{5.81}${\scriptsize$\pm.00$}
& $\mathbf{8.20}${\scriptsize$\pm.07$}
& $\mathbf{10.98}${\scriptsize$\pm.04$}
& $\mathbf{16.78}${\scriptsize$\pm.04$} \\
P3O
& $5.88${\scriptsize$\pm.01$}
& $8.41${\scriptsize$\pm.07$}
& $11.71${\scriptsize$\pm.17$}
& $18.47${\scriptsize$\pm.52$} \\
POMO
& $5.90${\scriptsize$\pm.00$}
& $8.48${\scriptsize$\pm.01$}
& $11.23${\scriptsize$\pm.04$}
& $16.91${\scriptsize$\pm.02$} \\
REINFORCE
& $5.89${\scriptsize$\pm.00$}
& $9.79${\scriptsize$\pm.00$}
& $11.25${\scriptsize$\pm.00$}
& $17.15${\scriptsize$\pm.01$} \\
PPO
& $5.90${\scriptsize$\pm.00$}
& $8.97${\scriptsize$\pm.06$}
& $11.81${\scriptsize$\pm.12$}
& $19.54${\scriptsize$\pm.95$} \\
\bottomrule
\end{tabular}
\end{table}

Table~\ref{tab:epoch} shows results at 100 training epochs. GRPO
achieves the lowest cost on all four benchmarks. However, this
comparison conflates algorithmic quality with compute: GRPO performs
$8\times$ more gradient updates per epoch than REINFORCE and POMO.
We address this in the next section.

\begin{figure}[t]
\centering
\includegraphics[width=0.95\textwidth]{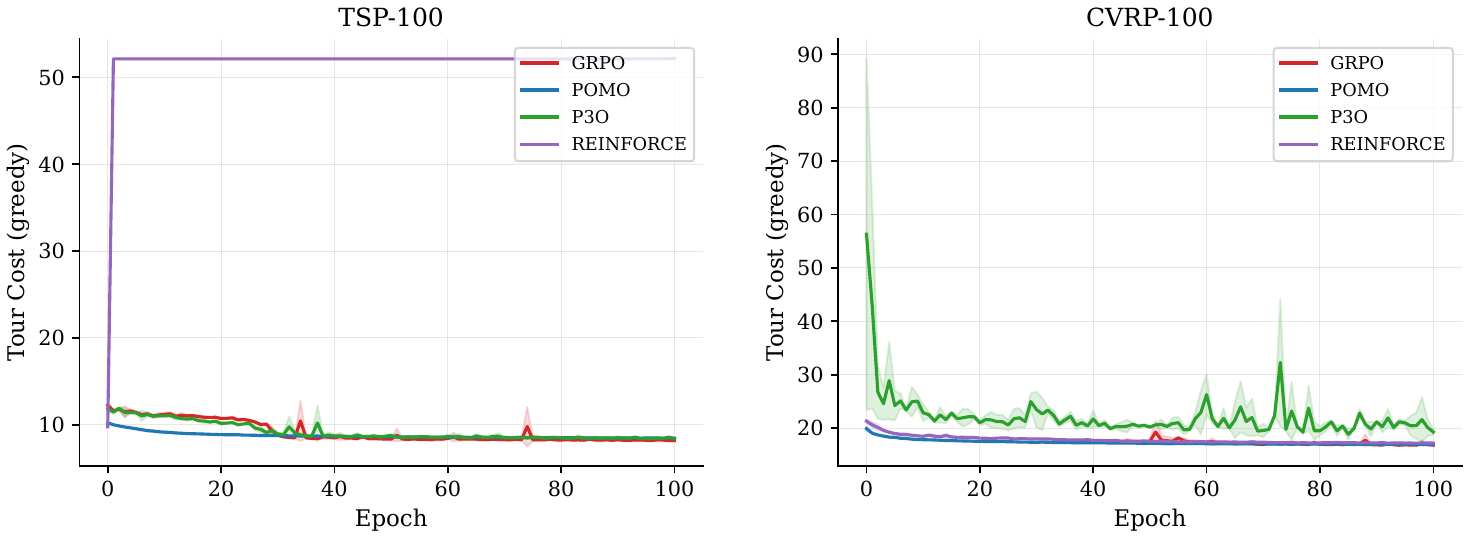}
\caption{Validation cost vs.\ training epoch on TSP-100 and
CVRP-100. REINFORCE (purple) learns during warmup (epoch~0)
but degrades once the rollout baseline activates on TSP-100.
GRPO (red) and POMO (blue) converge stably on both problems.
Shaded regions: $\pm$1 std over 2 seeds. PPO omitted for
clarity (see Table~\ref{tab:epoch}).}
\label{fig:curves}
\end{figure}

\subsection{Gradient-Update-Matched Comparison}\label{sec:compute_results}

To separate epoch-level progress from optimizer update budget, we evaluate
two additional settings on the largest instances (TSP-100 and
CVRP-100), where differences are most pronounced:

\begin{enumerate}
\item \textbf{Extended baselines}: POMO and REINFORCE trained for 800
epochs (${\sim}$312K gradient updates, matching GRPO at 100 epochs).
\item \textbf{Reduced GRPO}: GRPO at epoch 12 (${\sim}$39K gradient
updates, matching REINFORCE/POMO at 100 epochs), extracted from
training logs.
\end{enumerate}

\begin{table}[t]
\centering
\caption{Gradient-update-matched comparison on TSP-100 and CVRP-100. Gradient updates are matched across rows. Best per row in \textbf{bold}.}
\label{tab:compute}
\small
\setlength{\tabcolsep}{3pt}
\begin{tabular}{llccc}
\toprule
Updates & Method & Epochs & TSP-100 & CVRP-100 \\
\midrule
\multirow{3}{*}{$\sim$312K}
& POMO      & 800 & $\mathbf{8.07}$ & $\mathbf{16.51}$ \\
& GRPO      & 100 & $8.20$          & $16.78$ \\
& REINFORCE & 800 & $9.79^{\dagger}$ & $16.59$ \\
\midrule
\multirow{3}{*}{$\sim$39K}
& POMO      & 100 & $\mathbf{8.48}$ & $\mathbf{16.91}$ \\
& REINFORCE & 100 & $9.79^{\dagger}$ & $17.15$ \\
& GRPO      &  12 & $11.00$          & $21.09$ \\
\bottomrule
\multicolumn{5}{l}{\scriptsize $^{\dagger}$ Best checkpoint at
epoch 0; unchanged regardless of training duration.}
\end{tabular}
\end{table}

\begin{figure}[t]
\centering
\includegraphics[width=0.95\textwidth]{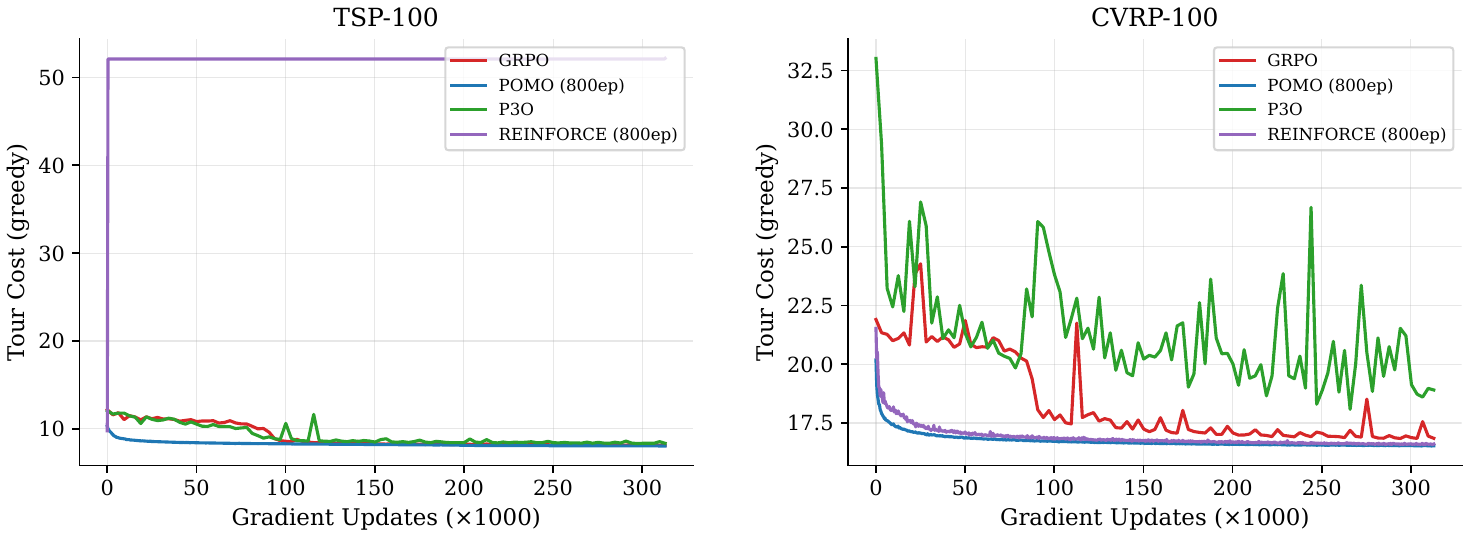}
\caption{Cost vs.\ gradient updates on TSP-100 and CVRP-100.
At matched gradient updates (${\sim}$312K updates), POMO at 800 epochs
(blue) slightly outperforms GRPO at 100 epochs (red).
REINFORCE at 800 epochs (purple) remains collapsed on
TSP-100 but improves on CVRP-100.}
\label{fig:compute}
\end{figure}

Table~\ref{tab:compute} reveals three key findings:
\textbf{POMO is the most efficient per gradient update.} At 312K
updates, POMO (800 epochs) slightly outperforms GRPO (100 epochs)
on both TSP-100 (1.5\% lower cost) and CVRP-100 (1.6\% lower cost). At 39K updates,
POMO (100 epochs) also leads. POMO's single REINFORCE update per
batch, combined with its multi-start symmetry exploitation, yields
the best quality per gradient step.

\textbf{GRPO converges in fewer epochs but not fewer updates.}
GRPO's epoch-level advantage (Table~\ref{tab:epoch}) reflects its
$8\times$ higher update rate, not a fundamentally faster learning
algorithm. At equal updates, GRPO is competitive with POMO (within
1.5--1.6\%) but does not surpass it.

\textbf{REINFORCE on TSP-100 fails structurally.} At 800 epochs
(312K updates), the best checkpoint remains at epoch~0 with cost
9.79, identical to the 100-epoch result. This confirms that the
failure is not a compute limitation but a structural instability of
the rollout baseline. On CVRP-100, by contrast, REINFORCE improves
to 16.59 at 800 epochs, outperforming GRPO at 100 epochs.

\subsection{Rollout Baseline Failure Analysis}\label{sec:failure}
Table~\ref{tab:stability} reports when each method reaches its best
TSP-100 validation checkpoint, distinguishing late-stage improvement from early plateauing or warmup-only behavior.
\begin{table}[t]
\centering
\caption{Best-checkpoint timing on TSP-100.}
\label{tab:stability}
\small
\begin{tabular}{lccc}
\toprule
Method & Best Epoch & Test Cost & Observed Pattern \\
\midrule
GRPO      & 97--99/100 & 8.20 & Late best checkpoint \\
POMO      & 94--98/100 & 8.48 & Late best checkpoint \\
P3O       & 94--95/100 & 8.41 & Late best checkpoint \\
PPO       & 9--18/100  & 8.97 & Early plateau \\
REINFORCE & 0/800      & 9.79 & Warmup-only best checkpoint \\
\bottomrule
\end{tabular}
\end{table}

The aggregate costs in Tables~\ref{tab:epoch} and~\ref{tab:compute}
hide an important temporal effect: several methods reach their best
checkpoint at very different stages of training. Table~\ref{tab:stability} summarizes the epoch at which each method attains its best TSP-100 validation performance, exposing whether the final result reflects stable late-stage convergence, early plateauing, or warmup-only learning.

In RL4CO, selecting the rollout baseline for the Attention Model wraps the baseline in a warmup mechanism. During epoch~0, this mechanism blends an exponential moving average estimate with the rollout baseline; from epoch~1 onward, it uses the pure rollout baseline.

During the warmup phase (epoch~0), the exponential baseline provides smooth gradient estimates, enabling rapid learning: the greedy tour cost drops from ${\sim}$52 (random policy) to ${\sim}$9.8 in a single epoch. When the rollout baseline activates at epoch~1, it uses this partially trained policy's greedy rollout as the new baseline.
However, the gradient estimates become noisy because the greedy
baseline is still far from optimal on TSP-100. This noise destabilizes
training: the policy degrades back to random performance (${\sim}$52),
which in turn degrades the baseline, creating a negative feedback loop
from which the system never recovers.

This failure is not a hyperparameter issue: it persists across both
random seeds and at 800 epochs ($8\times$ longer training). It is a
structural limitation of coupling the baseline to the policy on problems where the initial greedy policy is too poor to serve as a
useful reference.

GRPO sidesteps this entirely: with no external baseline, there is no feedback loop. The intra-group z-score normalization provides
informative gradients regardless of the absolute quality level, since even among poor trajectories, some are relatively better.

Fig.~\ref{fig:curves} and Fig.~\ref{fig:compute} visualize
these patterns. REINFORCE's cost
drops sharply at epoch~0 then reverts to random-level performance,
while GRPO and POMO show monotonic improvement.

\subsection{PPO and P3O}

PPO, which in RL4CO treats solution construction as a single-step
MDP with an episode-level learned critic~\cite{berto2025rl4co}, exhibits
early convergence followed by degradation on the larger instances, with best checkpoints at epochs 9--18 for TSP-100 and 5--24 for CVRP-100. The episode-level critic must predict total tour cost from the initial graph state alone, a task that becomes increasingly difficult as problem size grows.

P3O achieves the second-best cost on TSP at both sizes (5.88 on
TSP-50, 8.41 on TSP-100), but shows higher variability on CVRP-100
across the two tested seeds (inter-seed coefficient of variation 5.7\%, compared with 0.5\% for GRPO and 0.2\% for POMO). This suggests that the pairwise comparison mechanism may be more sensitive to noisy trajectory pairs in constrained routing settings.

% ==============================================================
\section{Discussion}\label{sec:discussion}
% ==============================================================

\textbf{When to use GRPO.}
GRPO is most valuable when the rollout baseline is fragile, on larger
instances, more constrained problems, or problems where greedy
heuristics perform poorly. For small instances like TSP-50 where all
methods converge to similar quality, the simpler REINFORCE or POMO
may be preferred. For practitioners who want baseline-free training without maintaining a frozen rollout policy or learned critic, GRPO provides a practical alternative.

\textbf{Sample efficiency vs.\ gradient efficiency.}
The gradient-update-matched comparison (Table~\ref{tab:compute}) equalizes gradient updates but not unique problem instances. Because RL4CO generates fresh instances each epoch, POMO at 800 epochs sees $8\times$ more unique problem instances (80M) than GRPO at 100 epochs (10M). GRPO's multi-trajectory group sampling and off-policy inner-loop reuse extract more learning signal per instance: GRPO achieves quality within 1.6\% of POMO while seeing $8\times$ fewer unique problems. POMO is more efficient per gradient update; GRPO is more efficient per unique instance. In domains where instance generation is expensive (e.g., simulation-based environments or real-world logistics), GRPO's lower instance requirement may be the deciding factor.

\textbf{Temperature annealing.}
GRPO and P3O use temperature annealing ($T{:}2{\to}1$) while
REINFORCE and POMO use $T{=}1$. This asymmetry is a potential
confound: higher initial temperature increases exploration diversity
in GRPO's group sampling. However, the REINFORCE failure on TSP-100
occurs during the exponential baseline warmup at $T{=}1$ and is
unrelated to the temperature schedule.

\textbf{Limitations.}
We evaluate only one architecture (6-layer AM) and two problem types. The conclusions about baseline fragility may not generalize to problems where greedy heuristics are already strong. All experiments use greedy decoding at inference;
search-based methods (beam search, sampling) may alter the ranking.

% ==============================================================
\section{Conclusion}\label{sec:conclusion}
% ==============================================================

We evaluated Group Relative Policy Optimization (GRPO) as a
baseline-free training algorithm for neural combinatorial optimization.
The central finding is that the rollout baseline, the standard
variance-reduction technique in NCO, introduces a structural
vulnerability on harder instances: on TSP-100, REINFORCE learns
during a brief warmup but degrades catastrophically once the rollout
baseline activates. GRPO avoids this failure entirely through
intra-group advantage normalization that requires no external
baseline.

At equal gradient updates, GRPO achieves solution quality within
2\% of POMO, a strong AM-based multi-start baseline, demonstrating that
baseline-free training is competitive with, though not superior to, baseline-dependent methods under a matched gradient-update budget. We also provide
the first evaluation of P3O for NCO, finding it competitive on TSP
but unstable on CVRP.

These results suggest that GRPO is a promising baseline-free alternative for NCO training, particularly when scaling to settings where baseline-dependent methods may become fragile. Future work includes extending the evaluation to larger instances (TSP-200+), additional problem types, and investigating whether GRPO's stability advantage compounds with curriculum learning or more complex environments.

Code and instructions to reproduce the experiments:
\url{https://github.com/carsepmo/grpo-nco}.

\bibliographystyle{IEEEtran}
\bibliography{reference}

\end{document}